# NETWORK SCIENCE

# A Bayesian machine scientist to aid in the solution of challenging scientific problems

Roger Guimerà[1,2]*, Ignasi Reichardt[2], Antoni Aguilar-Mogas[2,3], Francesco A. Massucci[2,4], Manuel Miranda[2], Jordi Pallarès[5], Marta Sales-Pardo[2]

Closed-form, interpretable mathematical models have been instrumental for advancing our understanding of the world; with the data revolution, we may now be in a position to uncover new such models for many systems from physics to the social sciences. However, to deal with increasing amounts of data, we need "machine scientists" that are able to extract these models automatically from data. Here, we introduce a Bayesian machine scientist, which establishes the plausibility of models using explicit approximations to the exact marginal posterior over models and establishes its prior expectations about models by learning from a large empirical corpus of mathematical expressions. It explores the space of models using Markov chain Monte Carlo. We show that this approach uncovers accurate models for synthetic and real data and provides out-of-sample predictions that are more accurate than those of existing approaches and of other nonparametric methods.

## INTRODUCTION

Since the scientific revolution, interpretable closed-form mathematical models have been instrumental for advancing our understanding of the world. Think, for example, of Newton's law of gravitation and how it has enabled us to predict astronomical phenomena with great accuracy and, perhaps more importantly, to interpret seemingly unrelated physical phenomena. With the data revolution, we may now be in a position to uncover closed-form, interpretable mathematical models of natural and even socioeconomic systems that were previously not amenable to quantitative analysis.

To be able to do this, however, we need to develop algorithms for automatically identifying these models (1–4). Following Evans and Rzhetsky, here, we call these algorithms, which would assist human scientists, "machine scientists" (2).

Attempts to design machine scientists date back to, at least, the 1970s and have led to very successful approaches in recent years. One of these approaches is based on genetic programming (5, 6). In this approach, closed-form mathematical expressions are represented as graphs, and, given a goodness-of-fit metric, populations of expressions are created and evolved in such a way that high-fitness expressions are selected for further exploration. Another successful approach is based on sparse regression (7–11). In this approach, closed-form mathematical models are assumed to be linear combinations of some (linear or nonlinear) "basis functions" of the independent variables, and sparse regression is used to select and weigh the relevant basis functions. This approach is particularly suited to learn differential equations (8–11), whose form often follows the assumption of linearity on relatively simple basis functions. For more general situations, sparse regression can be combined with genetic programs that automatically generate the basis functions, thus relaxing the need to know them a priori (12, 13).

Despite this remarkable progress, machine scientists have stumbled upon two major challenges. First, algorithms must balance goodness of fit and model complexity, thus avoiding overfitting. In general, this issue is dealt with by defining model complexity heuristically and then applying model selection criteria to the models that lay on the fit-complexity Pareto front. Both the definition of complexity and the choice of model-selection criterion are, however, hard to generalize and systematize. Second, machine scientists should, in principle, explore an arbitrarily large space of closed-form mathematical models. This is typically addressed by using methods such as genetic programming, which have no guarantees of actually exploring the best models more frequently; or by restricting the search space, for example, to linear combinations of the basis functions in sparse regression approaches, thus leaving out potentially valid models.

Here, we propose a Bayesian approach to the definition of machine scientists. To address the fit-complexity trade-off, we obtain the posterior probability of each expression from basic probabilistic arguments and explicit approximations. The posterior, which leads to consistent model selection, naturally combines the goodness of fit and a prior over expressions that accounts for model complexity. To establish the prior over expressions, we compile a corpus of closed-form mathematical models from Wikipedia and then use a maximum entropy approach to formalize a prior that is statistically consistent with the corpus (14). To address the challenges related to the exploration of the space of closed-form mathematical expressions, we introduce a Markov chain Monte Carlo (MCMC) algorithm that samples from the posterior over expressions. We demonstrate that the Bayesian machine scientist successfully recovers the true generating model when fed with synthetic data, even in situations in which state-of-the-art machine scientists fail (4). We also demonstrate that the machine is able to uncover accurate, closed-form mathematical models for systems for which no closed-form model has agreed on. Last, we find that the machine scientist provides out-of-sample predictions that are more accurate than those of other machine scientists and of standard nonparametric machine learning approaches, such as Gaussian processes (15).

## RESULTS

### Bayesian formulation of the problem and expression plausibility

Let us first formalize the problem in probabilistic terms. Consider a property $y$ that can be expressed as an unknown, closed-form mathematical function $y = F(x, \theta)$ of $K$ variables $x = \{x_1, ..., x_K\}$ and $L$ parameters $\theta \in R^L$ [for example, $y = \sin(\theta_1 x_1)$ or $y = \theta_1 x_1 + \theta_2 x_2$]. Given some data $D = \{(y^1, x^1), ..., (y^N, x^N)\}$, and assuming that the measurements have some experimental error $y^k = F(x^k, \theta) + \epsilon^k$, the Bayesian machine scientist assigns to each possible closed-form mathematical expression $f_i$ a plausibility $p(f_i|D)$ given by the marginal posterior

$$p(f_i|D) = \frac{1}{Z}\int_{\Theta_i} d\theta_i\, p(D|f_i, \theta_i)\, p(\theta_i|f_i)\, p(f_i) = \frac{\exp[-\mathcal{L}(f_i)]}{Z} \quad (1)$$

where $\theta_i$ are the parameters associated with expression $f_i$, the integral is over the space $\Theta_i$ of possible values of these parameters, $Z = p(D)$ does not depend on $f_i$, and $p(f_i)$ is the prior over expressions. The quantity

$$\mathcal{L}(f_i) \equiv -\log[p(D, f_i)]$$
$$= -\log\left[\int_{\Theta_i} d\theta_i\, p(D|f_i, \theta_i)\, p(\theta_i|f_i)\, p(f_i)\right] \quad (2)$$

is the description length of model $f_i$, that is, the number of nats needed to jointly encode the data and the model with an optimal code (16). Although in general the description length cannot be calculated exactly, it can be approximated in a number of ways (17, 18); here, we take one of the simplest approximations

$$\mathcal{L}(f_i) \approx \frac{B(f_i)}{2} - \log p(f_i) \quad (3)$$

where $B(f_i)$ is the Bayesian information criterion (BIC) of expression $f_i$ and can be readily calculated from the data (Supplementary text S1) (11, 17). This is a first-order approximation to the description length and holds when the likelihood $p(D|f_i, \theta_i)$ is peaked around the maximum likelihood parameters $\theta_i^*$ and the prior is smooth in this region; when the number of experimental points is small, the approximation may fail, and one would need to get more precise estimates of the description length such as the generalized BIC (19) or even calculate numerically the integral over the parameters. Beyond these potential limitations, Eqs. 1 to 3 naturally combine the goodness of fit of a model and its structural complexity, which is captured by the prior over expressions. In addition, in the limit of large datasets, we have $|B(f_i)| \gg |\log p(f_i)|$; since $B(f_i)$ is consistent, the machine scientist is also consistent, that is, in this limit it prefers the correct expression over any other expression with probability approaching 1.

[1]ICREA, Barcelona 08010, Catalonia, Spain. [2]Department of Chemical Engineering, Universitat Rovira i Virgili, Tarragona 43007, Catalonia, Spain. [3]Division of Research, Economic Development and Engagement, East Carolina University, Greenville, NC 27858, USA. [4]SIRIS Lab, Research Division of SIRIS Academic, Barcelona 08003, Catalonia, Spain. [5]Department of Mechanical Engineering, Universitat Rovira i Virgili, Tarragona 43007, Catalonia, Spain.
*Corresponding author. Email: roger.guimera@urv.cat

**Sampling from the posterior distribution over expressions**
The Bayesian machine scientist explores the space of closed-form mathematical expressions using MCMC. In particular, we introduce three move types that enable one to go from any closed-form expression to any other closed-form expression, thus enabling the machine scientist to explore, given enough time, the whole space of closed-form mathematical expressions (Materials and Methods) (Fig. 1). Regardless of the frequency of each move and other parameters of the Markov chain, MCMC samples expressions $f_i$ from the stationary distribution $p(f_i|D)$ (fig. S1). Although MCMC is slower than some alternative machine scientists that put emphasis on speed, such as evolutionary feature synthesis (EFS), the only dependency on the number of data points is in the estimation of the BIC of each model, and therefore its complexity scales as any other method using least squares to fit model parameters.

For model selection, the Bayesian machine scientist can use the most plausible expression from an MCMC run, that is, the maximum a posteriori (or minimum description length) expression. However, MCMC naturally samples over the whole space of models, thus generating arbitrarily long sequences of expressions; as we show below, this leads to a more complete characterization of the expression space and to higher out-of-sample prediction accuracy.

**Estimation of prior probabilities using a corpus of mathematical expressions**
For the Bayesian machine scientist to be able to estimate the plausibility of a given expression, it needs to estimate the prior probabilities $p(f_i)$. A common approach in model selection is to have no a priori preference for any given model over the others and assume that $p(f_i)$ is the same for all models (17, 18). In that case, and within our approximation for the description length, the most plausible model is simply the one with the lowest BIC. This is a consistent approach and generally reasonable when comparing a small number of simple models. However, it is inappropriate when considering a finite dataset and a very large (potentially infinite) space of mathematical expressions because one can always find a very complex model that fits the data arbitrarily well even with very few parameters. These unnecessarily complex models are likely to generalize very poorly in the same way that models with many parameters do—this structural overfitting is thus akin to traditional overfitting but arises from the large number of mathematical models considered rather than the large number of parameters (Supplementary text S4 and fig. S5). From this perspective, the prior over expressions acts as a model regularizer. It would also be possible to add other regularizers to expression trees, but, within our Bayesian framework, tree regularizers could still be cast as non-uniform priors, although they may be formally more complex and harder to interpret than the ones that we introduce below.

Given these considerations, the machine scientist needs to "learn" the a priori plausibility of models. Human scientists obtain this prior knowledge by studying science books and becoming familiar with the mathematical expressions that appear in them. We take a similar empirical approach to define the prior expectations of the machine scientist—we compiled a corpus of 4080 mathematical expressions that are included in Wikipedia entries (Materials and Methods)—and use these expressions to shape the prior expectations of the machine scientist, that is, to establish the statistical properties expected a priori for expressions (Fig. 2). To do this, we use an approach based on exponential random graphs (20–23). In particular, we choose a prior that generates expressions with the same average number of each type of operation [and their squares (23, 24)] as in the empirical corpus and, given this constraint, satisfies the maximum entropy principle (14, 23) (Materials and Methods; Supplementary text S2, fig. S2, and tables S3 and S4). As we show in Fig. 2, this prior generates mathematical expressions statistically consistent with the corpus. Note that the Bayesian machine scientist is not restricted to the 4080 expressions in the corpus, or even to arbitrary combinations of these expressions—all closed-form mathematical expressions are valid and can be visited by the MCMC; those that are statistically similar to the corpus are simply more plausible a priori. Note also that, as mentioned earlier, for large amounts of data, the prior washes out and the description length, as approximated in Eq. 3, is consistent regardless of the selection of prior.

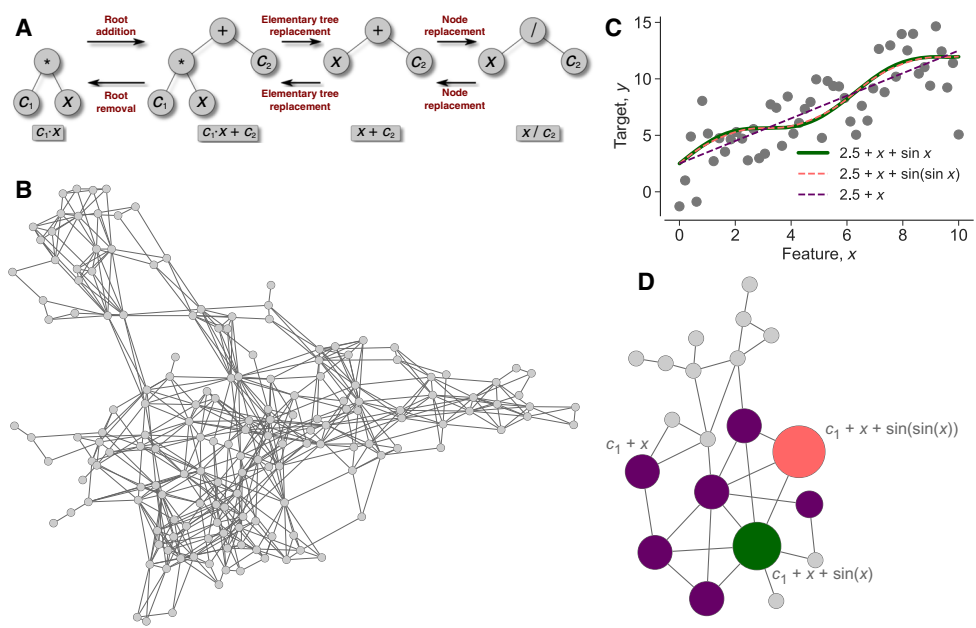

**Fig. 1. MCMC to systematically explore the space of closed-form mathematical expressions.** (**A**) Closed-form mathematical expressions can be represented as trees, whose internal nodes are functions (operations) and whose leaves are variables and parameters. To explore the space of mathematical expressions, we implemented three move types (Materials and Methods): (i) root addition/removal—we added/removed the root of the tree; (ii) elementary tree replacement (we define an elementary tree as a subtree containing, at most, one function)—we replaced one elementary tree by another (for example, $c_1 x$ by $x$); (iii) node replacement—we replaced one node (function, variable, or parameter) by another. Each of these moves was proposed with a certain frequency and accepted or rejected using the Metropolis' rule (Materials and Methods) (*36*). (**B**) To validate the MCMC, we explored the space of all expressions with one variable, one parameter and using only the functions {+, sin}. For this experiment only, we restricted the size of the expression tree to be smaller than eight nodes, and we assume that all expressions are equally plausible a priori, $p(f_i) = $ const. Each node in the network represents a different expression, and links represent jumps from one expression to another resulting from the MCMC moves described in (A). The size of each node is proportional to the number of times they are visited by the MCMC. In the absence of data, the MCMC explores all possible expressions with equal probability. (**C**) We generated synthetic data with the expression $y = a + x + \sin(x)$, to which we added noise (circles). We show how the correct expression and two closely related expressions fit the data. (**D**) We explored the same space as in (B) but considering the synthetic data in (C). When data are taken into account, expressions that are more plausible given the data are visited more frequently (larger nodes), and many expressions are too implausible and, therefore, not visited at all (fig. S1). The color of each node corresponds to the color of the curves in (C).

### Validation of the Bayesian machine scientist with synthetic data

Having addressed the methodological challenges in the definition of the Bayesian machine scientist, we next demonstrate the ways in which it can be used and illustrate how one can get insights by using it. First, we test that, when fed with synthetic data, the machine scientist recovers the models that truly generated the data. We start by selecting an arbitrary expression and setting its parameters to values uniformly selected from [−2, 2]. The selected expression (Fig. 3) is $F(x_1, x_2; \theta_1, \theta_2) = x_1(\theta_1 + x_2) \cos(x_1)/[\theta_2 \log(\theta_2)]$, with $\theta_1 = -1.19$ and $\theta_2 = 0.29$. Note that this expression is not one of the expressions present in our empirical corpus.

We then feed the machine scientist with 400 noisy data points generated using this expression. The Bayesian machine scientist finds the correct model and assigns to it the maximum plausibility or, equivalently, the minimum description length (Fig. 3). To evaluate to which extent it is remarkable that the Bayesian machine scientist finds the correct model in this situation, we attempt the same task with state-of-the-art machine scientists (*4*), including two methods based on genetic programing [Eureqa (*5*) and ϵ-lexicase selection (EPLEX) (*6*)] and a method that combines genetic programming with sparse regression (Materials and Methods) [EFS; (*13*)]. We find that, from the same dataset, none of these methods are able to recover the correct model and that they tend to structurally overfit the data (Supplementary text S3).

The ability of the Bayesian machine scientist (and of any other method) to identify the correct model requires a minimum amount of data points. We find that, for this synthetic dataset, and even though the other approaches do not identify the correct expression with 400 points, the Bayesian machine recovers the true model with as few as 100 points (Supplementary text S3 and fig. S3).

Next, we investigate whether the machine is able to recover the differential equations that govern the Rössler system (*25*), and what is the effect of increasing observational noise in the equation discovery process. Since we are mostly interested in the effect of increasing noise in the target variable, we assume that the only measurement error is in the derivatives, although in some real situations derivatives may need to be estimated numerically from noisy measurements of the variables. In those situations, it may be necessary to use advanced techniques to estimate the derivatives (*26*). Under the conditions of our experiment, we find that the machine is able to recover the correct differential equations when the derivatives are measured with moderate noise (Fig. 4). When the measurement noise is high, the true expressions are still regarded as very plausible, but the machine identifies as the most plausible ones expressions that are "regularized" versions of the exact models. In these regularized models, small terms are disregarded; in all cases, the most plausible models are almost indistinguishable from the true ones (fig. S4). Thus, as expected, the machine scientist automatically adjusts the complexity of the models to the quality of the data (Supplementary text S4). Again, these results stand in contrast to those of alternative machine scientists [with the exception of pure sparse regression methods particularly suited to reverse-engineer differential equations (8–11), which would be able to recover the true expressions, at least in the case with low noise]. Even for the simplest case in this experiment (inference of $\dot{x}$ with a low level of noise), all benchmark machine scientists fail to recover the true model and tend to overfit structurally (Supplementary text S4).

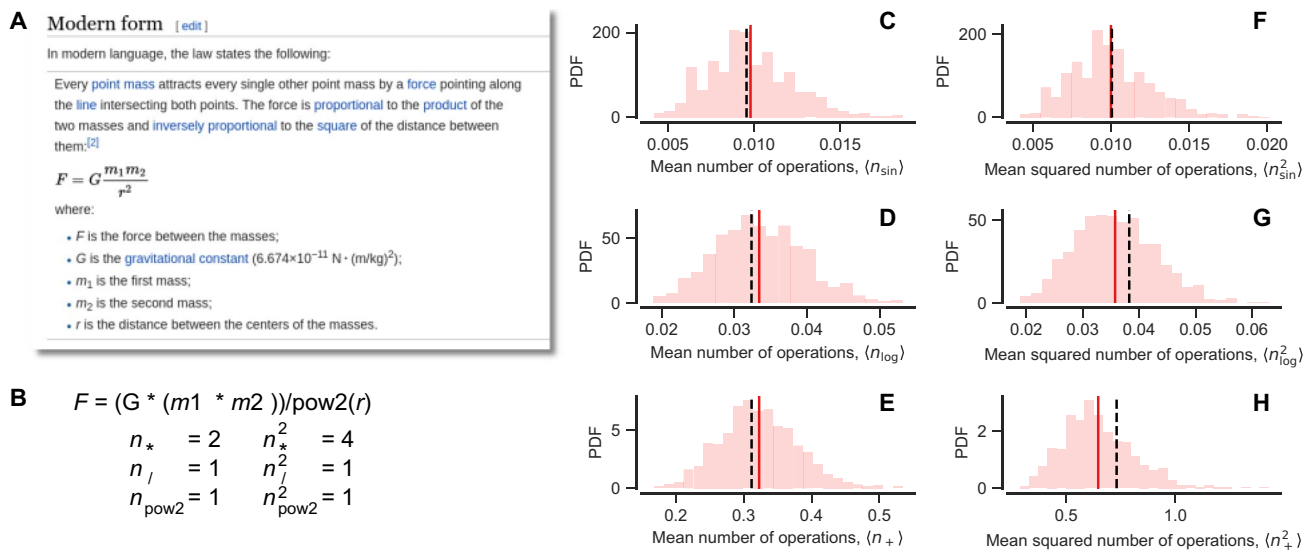

**Fig. 2. The prior probability distribution generates plausible expressions.** (**A**) We extracted 4080 mathematical expressions from Wikipedia pages listed under the category "List of scientific equations named after people", for example, the page devoted to "Newton's law of universal gravitation" (see also table S2). (**B**) After parsing each expression (Materials and Methods), we counted the number of each operation in the expression and the square of these numbers. In Newton's law of gravitation, we counted two products, one division, and one square. The complete list of operations is listed in table S1. Then, we parametrized the prior distributions $p(f_i)$ and fit the parameters so as to generate expressions that have realistic numbers (and squares) of each operation (Materials and Methods). (**C** to **H**) We ran the MCMC to draw sets of 4080 synthetic expressions from the prior distribution. We compared these sets of 4080 synthetic expressions with the 4080 empirical expressions obtained from Wikipedia. In particular, for each set of 4080 expressions, we counted the mean number $\langle n_o \rangle$ of each operation $o$ per expression [(C) sine, (D) logarithm, (E) sum], or the mean $\langle n_o^2 \rangle$ of the square of those numbers (**F** to **H**) and plot their probability density function (PDF; pink). The red vertical line indicates the mean of the distribution, which is close to the empirical observation, represented by a dashed black line. The empirical value is always well within the expected ranges of the expressions generated by the MCMC. The results in the figure correspond to expressions with up to five variables and eight parameters.

Last, we test the behavior of the machine scientist when presented with data that are generated from a model that does not have a closed-form mathematical expression in terms of the basic functions that it uses. In particular, we generate synthetic data using Bessel functions $J_\alpha(x)$ with $\alpha \in \{0, 1, 2, 3, 4\}$. We find that the machine scientist is able to identify closed-form expressions that are as accurate as high-order Taylor expansions of the exact functions. We also find that, when the synthetic data are noisy and the machine scientist is restricted to choose among low-order series expansions, it consecutively chooses first-order, second-order, or third-order expansions, as the range of observed data increases and as the noise decreases, as one would expect (Supplementary text S5 and fig. S6).

### Use of the machine scientist on small datasets and on the Nikuradse dataset

Next, we turn to the analysis of real datasets. First, we analyze how the machine scientist provides insights into problems for which there are scarce and noisy data. We focus on three datasets that have been the subject of recent analyses and for which models have been proposed: a dataset on funding success in different European countries (27); a dataset on cell-to-cell stresses (28); and a dataset on stocks of salmon in the Fraser River in British Columbia, Canada (29).

For each of these three datasets, we compare existing models to models identified by the machine scientist in terms of their plausibility $p(f_i|D)$, their BIC $B(f_i)$, and their cross-validation error (Supplementary text S6 and tables S5 to S7). In all cases, the machine scientist identifies at least one model that is better than existing models in a Pareto sense, namely, better in at least one of the three performance metrics without being worse in any of the others.

Last, we show in more detail how the machine scientist can help us to solve major open scientific problems. For this, we focus on the classical experiment of turbulent friction in rough pipes performed in the early 1930s by Johann Nikuradse (30–33). In his experiments, Nikuradse measured the turbulent friction $\lambda$ as a function of the roughness $x_D$ of the pipe and the Reynolds number $x_R$. We take the original Nikuradse dataset and feed it to the machine scientist to study possible analytical expressions for the turbulent friction. A typical MCMC run uncovers numerous expressions that fit all observed data remarkably well (Fig. 5, A and B). We compare these expressions to the best expressions uncovered by other machine scientists (Materials and Methods): Eureqa (5), EFS (13), and EPLEX (6) (see Supplementary text S7 and fig. S10 for EPLEX). The expressions uncovered by the Bayesian machine scientist fit the Nikuradse dataset better than those uncovered by Eureqa, which, in turn, is significantly better than all other benchmark methods.

The Bayesian machine scientist does not find any candidate expression that is overwhelmingly more plausible than all the others;

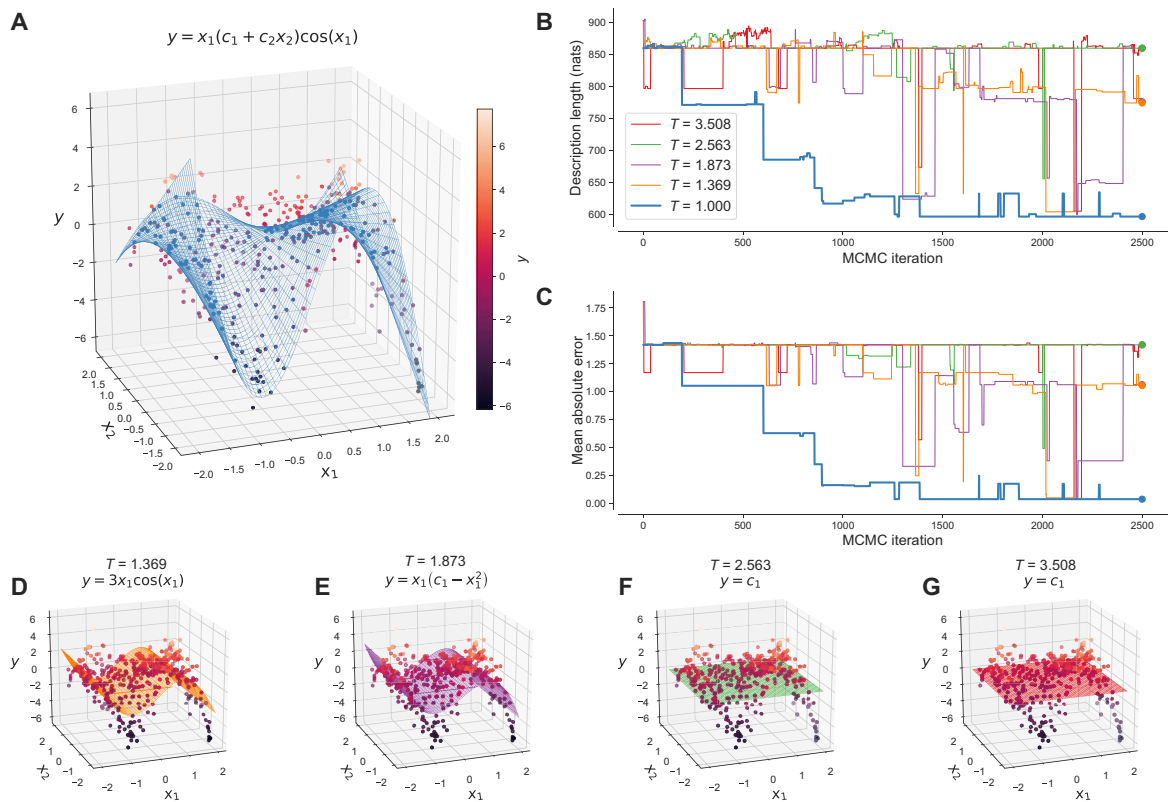

**Fig. 3. Recovery of an arbitrary expression from synthetic data.** We drew an expression with two variables ($x_1$ and $x_2$) and two parameters ($\theta_1$ and $\theta_2$) from the prior distribution $p(f_i)$; the drawn function is $F(x_1,x_2;\theta_1,\theta_2) = x_1(\theta_1 + x_2)\cos(x_1)/[\theta_2 \log(\theta_2)]$, with $\theta_1 = -1.19$ and $\theta_2 = 0.29$. With this function, we generated 400 synthetic points $y^k = F(x_1^k,x_2^k;\theta_1,\theta_2) + \varepsilon^k$, with $(x_1^k,x_2^k)$ uniformly chosen in $[-2,2]^2$, and $\varepsilon \sim N(0,1)$ normally distributed [points in (**A**) and (**D**) to (**G**); the color of the points corresponds to the value of $y$ as indicated by the color bar]. We fed these synthetic data to the Bayesian machine scientist and run 2500 steps of the MCMC algorithm. (**A**) After 2500 steps, the most plausible model identified (blue surface) is $f(x_1,x_2;\theta_1,\theta_2) = x_1(c_1 + c_2 x_2)\cos x_1$, which coincides with the correct model up to indeterminacies in the dependency on the parameters that cannot be resolved without varying their values (see movie S1 for the complete evolution of the MCMC). (**B**) The blue line indicates the evolution of the description length $L$ during the MCMC (up to an irrelevant additive constant that comes from the prior normalization and therefore affects all points equally). Lower description lengths correspond to more plausible models. The MCMC starts from an expression with high description length, but after 1000 to 1500 steps, it equilibrates and samples from the stationary distribution $p(f_i|D)$. (**C**) The blue line indicates the evolution of the mean absolute error of each model sampled by the MCMC with respect to the true model. The error is calculated over a grid in $[-2,2]^2$, that is, not at the observed points. (**D** to **G**) The MCMC is prone to getting trapped in local maxima of the plausibility (local minima of the description length). To explore the expression space more efficiently, we used parallel tempering (*38*), a technique used in statistical physics to study disordered systems with rugged energy landscapes (Materials and Methods). Besides the main MCMC in (A), parallel tempering keeps a number of parallel MCMCs that sample the distributions $p(f_i,T_k) \propto \exp[-B(f_i)/2T_k + \log p(f_i)]$ parametrized by "temperatures" $T_k$ (Materials and Methods). In our case, higher temperatures correspond to simpler expressions. At each MCMC step, we attempted a swap of expressions at contiguous $T_k$ using Metropolis' rule. The evolution of the description length and the mean absolute error of the parallel MCMCs are represented in (B) and (C) with the same colors as in (D) to (G).

rather, it uncovers a collection of similarly plausible models. This has two important implications. First, it points toward the need to revisit our tendency to look for single "best models" from data. Second, it suggests that, when using the machine scientist to make predictions, we should average over the whole ensemble of plausible models (*34*, *35*). In particular, the posterior predictive distribution for a point $y^k$ can be approximated as

$$p(y^k|D;x^k) \approx \sum_i \delta\left(y^k - f_i(x^k,\theta_i^*)\right) p(f_i|D) \quad (4)$$

where $\theta_i^*$ is the maximum likelihood estimator of $f_i$'s parameters, $\delta(x)$ is the Dirac delta function, and the sum runs over all possible expressions $f_i$. Note that estimating the posterior predictive distribution in this way makes interpretation of the predictions harder—even if each model $f_i$ is interpretable, averaging leads to a noninterpretable effective model. However, it is important to point out that this is the most comprehensive approach possible, in the sense that, even if one is certain that the data were generated with a single, unknown model $F$, the best predictive distribution comes from taking all models into consideration, each weighted by its plausibility $p(f_i|D)$ (*34*).

In practice, the average over models can be computed using the MCMC sample collected by the machine scientist, and we can use the median of the posterior distribution $p(y^k|D;x^k)$ to make predictions for $y^k$ that minimize the mean absolute predictive error. By considering, among all sampled models, the one that most resembles this median prediction, we obtain a single closed-form mathematical model that is optimally predictive—we call this model the median predictive model. To test the predictive power of the median predictive model in the Nikuradse dataset, we compare it to the alternative machine scientists (Eureqa, EPLEX, and EFS), as well as to a standard non-parametric Bayesian approach, Gaussian processes (15). In particular, we test the ability of all approaches to generalize to data never seen before (Fig. 5, C and D). We find that the predictions of the Bayesian machine scientist are significantly more accurate than those of all alternative approaches.

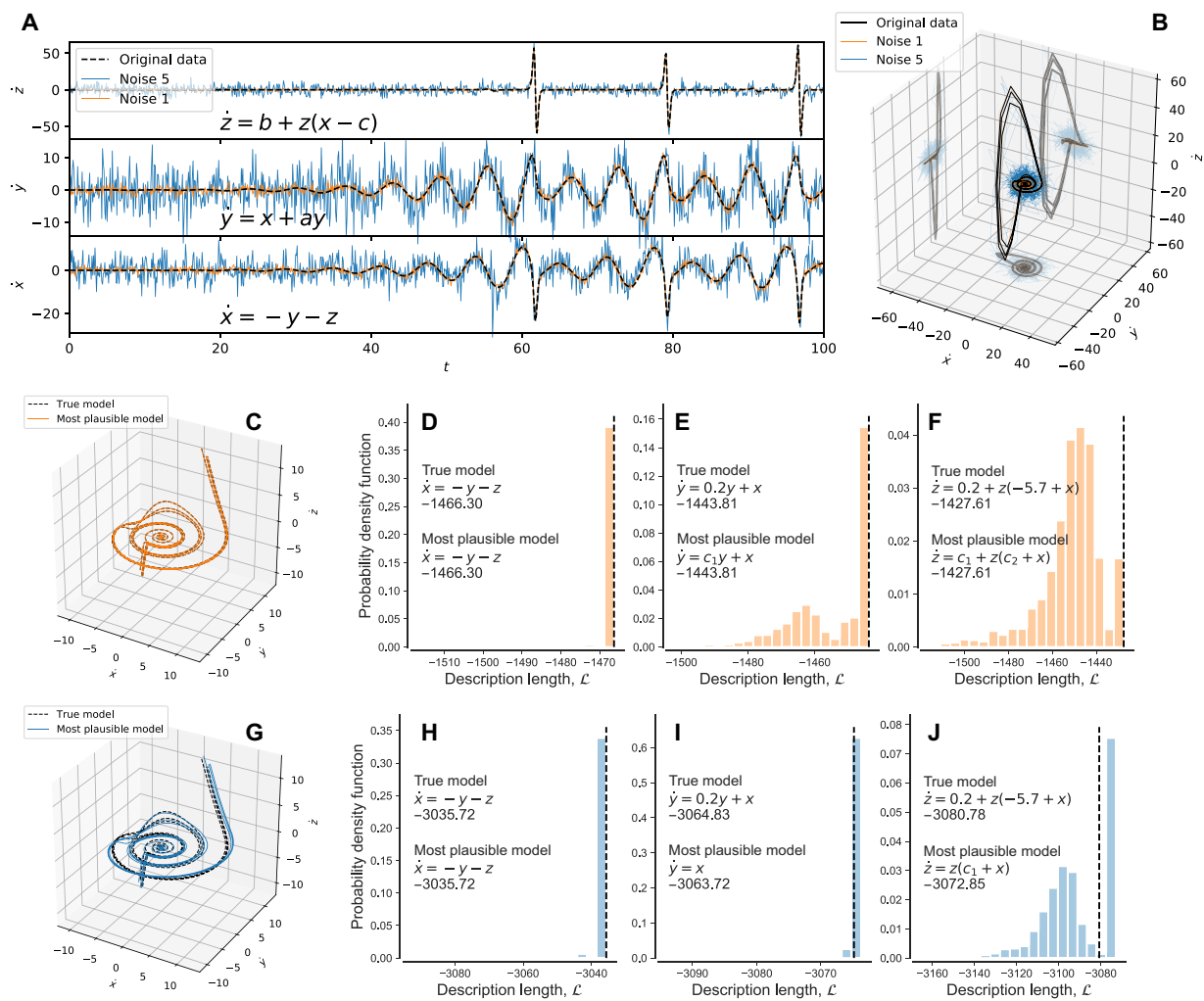

**Fig. 4. Recovery of the differential equations governing the Rössler system.** (**A**) We generated synthetic data for the Rössler system (equations shown in the panels) with $a = b = 0.2$ and $c = 5.7$. We assume that $(x, y, z)$ are measured without error, but the functions that we aim to recover $(\dot{x}, \dot{y}, \dot{z})$ are measured with two levels of Gaussian noise [e.g., $\dot{x}^k = F(x^k, y^k, z^k) + \varepsilon^k$ with $\varepsilon^k \sim N(0, \sigma_\varepsilon)$]: orange, $\sigma_\varepsilon = 1$; blue, $\sigma_\varepsilon = 5$. (**B**) Trajectories in velocity space $(\dot{x}, \dot{y}, \dot{z})$ for the levels of noise that we considered. (**C** to **J**) For each level of noise, we sampled 10,000 models for $\dot{x}$ using the machine scientist and the same for $\dot{y}$ and $\dot{z}$. (C and G), For each level of noise, we plot the most plausible model obtained. (D to F and H to J) We plot the distribution of description lengths (up to an irrelevant additive constant) for the models sampled. (D to F) For low noise, the most plausible model among those sampled always coincides with the true model, as indicated by the dashed vertical line. (H to J) For high noise, $\dot{x}$ was recovered exactly, whereas for $\dot{y}$ and $\dot{z}$, the most plausible models corresponded to regularized versions of the true models, in which terms that are comparatively small because of the factor 0.2 are dropped. In all cases, the true model (dashed vertical line) is, at least, among the most plausible ones.

The median predictive model, being a function of the roughness only for large Reynolds numbers, also predicts the expected limiting scaling for the turbulent friction (Fig. 5, E and F), which is remarkable considering that (i) most of the observed data correspond to a regime with different physics (31) and (ii) many of the sampled models do not scale correctly (fig. S9). Last, to fully exploit the potential of the machine scientist to obtain interpretable models of turbulent friction, we use it in combination with the known physics of the problem. In 1933, Prandtl suggested that the function $\tilde{\lambda} = (100\lambda)^{-1/2} - 2\log D/r$ should be a universal function depending only on the so-called roughness Reynolds number $k_s^+(\mathrm{Re}, r/D)$ (30, 33). We use the machine scientist to obtain the most plausible form for such universal function and get $\tilde{\lambda} = 1.73 + 0.64 \times 0.96^{k_s^+} - 2.62 \times 0.64^{k_s^+}$ (Fig. 5G); this very simple model on the scaled variable does indeed provide an excellent fit to the original (unscaled) data (Fig. 5H).

## DISCUSSION

In the age of data, there is a pressing need to design algorithms capable of helping in the scientific process, from assisting in the proof of theorems to parsing scientific texts. Discovering closed-form mathematical models was one of the earliest tasks attempted, and yet, despite much progress in the area, two related difficulties have arisen repeatedly—deciding what is the correct balance between goodness of fit and model complexity and searching systematically through the space of models.

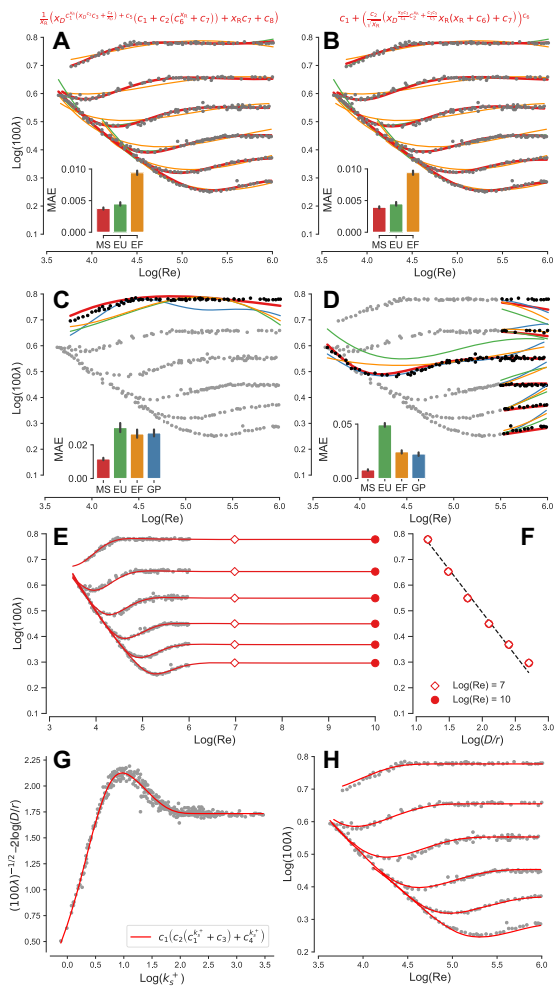

**Fig. 5. The machine scientist gives accurate closed-form models and out-of-sample predictions for the Nikuradse dataset.** The dataset contains measurements of the turbulent friction $\lambda$ in rough pipes as a function of the Reynolds number $x_R$ and the inverse of the relative roughness $x_D$. (**A** and **B**) We fed the machine scientist with the Nikuradse dataset (gray) and sample more than 18,000 models using MCMC as described in the text. The two graphs show the fit of two of the most plausible models (red), indicated at the top of each panel. For comparison, we show the best models identified by alternative machine scientists Eureqa (green) and EFS (orange). The inset in each graph shows the mean absolute error (MAE) of the machine scientist (MS), Eureqa (EU), and EFS (EF) (Materials and Methods), and error bars indicate the 95% confidence interval for the mean. (**C** and **D**) We fed the machine scientist with the data points in gray and sample 20,000 models using MCMC. We made predictions (red) for the unobserved data (black) using the median predictive model (see text). For comparison, we show the predictions of Eureqa, EFS, and Gaussian processes (GP) (Materials and Methods). The inset in each graph shows the mean absolute error for all approaches. (**E**) We considered, again, the models obtained from the whole data as in (A) and (B) but asked what the median predictive model predicts in the fully turbulent regime, well above the highest observed Reynold number (31). Despite the fact that individual sampled models have different behaviors in this region (fig. S5), the median predictive model [obtained as in (C) and (D)] predicts that turbulent friction stays constant at large Reynolds numbers, in agreement with theory (31–33). (**F**) The machine scientist also predicts the correct scaling $(D/r)^{-1/3}$ for the turbulent friction in this regime (dashed line). (**G**) Scaling of the Nikuradse dataset proposed by Prandtl (with $k_s^+(\text{Re}, r/D) = \text{Re} \times \sqrt{\lambda/32} \times D/r$) (30, 33) and most plausible model identified by the machine scientist for the universal scaling function. (**H**) When unscaled, this simple universal function provides a good description of all the data.

These are fundamental difficulties that stem from the nature of the problem that we aim to solve, and, therefore, no approach can possibly avoid them completely—one cannot establish an assumption-free measure of model complexity that serves all purposes or make the search space small without risking to leave out valid models or even the true generating model.

However, the Bayesian approach that we have introduced here forces us to be explicit and transparent about all the necessary assumptions and approximations to deal with these challenges. With regard to the fit-complexity trade-off, model complexity enters through our prior expectations about expressions. The use of an empirical corpus is somewhat arbitrary, and one may need to tune the corpus to the problem at hand; but it is reasonable to expect that, lacking any data, the machine scientist should propose models that statistically resemble existing phenomenological models in the literature. With our maximum entropy approach, we guarantee that the prior distribution is the most agnostic distribution satisfying this constraint. Besides, our approach makes it clear that, given enough data, prior expectations become irrelevant, and the machine scientist will consistently prefer the true model with probability approaching 1. With regard to the search through the space of models, MCMC guarantees that, asymptotically (that is, given enough data and time), the Bayesian machine scientist samples from the posterior distribution and, because it is consistent, visits the true model with the highest frequency.

Despite these advantages, our approach is not, of course, exempt of difficulties, the most pressing of which we have already pointed out. First, despite the virtues of MCMC for sampling the expression space, it may be necessary to develop more efficient approaches for very large datasets, without losing the guarantees provided by MCMC. Evidence for the need for these approaches would come from the observation that different MCMC runs on the same dataset result in different description length distributions, which would indicate that the stationary posterior is not being sampled correctly. Alternatives to sampling the full posterior using MCMC may range from Bayesian inference methods such as variational approximations to mathematical optimization methods, if suitable representations of the problem can be found and we are content with obtaining a single model and some bounds to its optimality. Second, it may be desirable to develop and compare other approaches to setting priors for expressions, perhaps based on noninformative priors or symmetry considerations, or empirical approaches different from the one that we have adopted here. A situation in which this need would be obvious is when the proposed models clearly violate some basic features of the desired solution, such as some particular limiting behavior or dimensional consistency. If such additional information is available, then it would be appropriate to formalize it in the prior. Last, in some situations involving small datasets and leading to broad likelihoods in parameter space, it may be necessary to use approximations to the description length that are more accurate than those based on the BIC, for example, the generalized BIC (18). However, this problem would be more difficult to diagnose in a practical situation because, even for a single dataset, the BIC can be a good approximation for some models but not for others. In any case, situations in which the BIC is a poor approximation are likely to involve small or very noisy datasets, situations in which the importance of the BIC is relatively small compared to that of the prior and simple models will be sampled anyway.

These challenges notwithstanding, our results suggest that, in practice, the Bayesian machine scientist is able to uncover models that are good in terms of both describing the observed data and predicting new data. Our approach can also be used in other contexts, for example when a phenomenological model has been proposed from data, and we aim to find whether there are other models that are more plausible or, at least, similarly plausible. Or, given two conflicting theories for the same process, the Bayesian machine scientist could be used to establish which one is more plausible and to what extent, and whether there exist other reasonable models. From a broader perspective, our approach sets the basis for further developing theories of model discoverability: Is it always possible to identify the correct model given the data? And if not, then under which conditions is the true model discoverable? In addition, our approach opens the door to addressing fundamental questions related to the limits of predictability for mathematical models: To what extent can a system be accurately predicted? Are closed-form, interpretable models as expressive as machine learning models such as deep neural networks? And if so, why? These are all important questions whose answer may lead to significant new insights about the scientific process and about our capacity to understand and describe the world using mathematics.

## MATERIALS AND METHODS
### MCMC moves for mathematical expression sampling
To perform a systematic exploration of the space of closed-form mathematical expressions, we represented closed-form mathematical expressions as expression trees. In these trees, internal nodes represent operations (for example, sum or exponential) and leaves represent variables or parameters (Fig. 1). To avoid problems with improper priors, we limited the size of expressions trees to 50 nodes. Although, in principle, one could make this number arbitrarily large, all parameter values and results in the paper are obtained using this value.

We classified the nodes of the expression tree based on the number of offspring that they have. Leaves (variables and parameters) have no offspring, whereas operations can have one offspring (operations that take only one argument, like the exponential function) or two offspring (operations that take two arguments, such as the sum). In table S1, we list all the operations that the machine scientist uses for building models.

An elementary tree (ET) is an expression tree (thus, a mathematical expression) that contains at most one operation. For the implementation of the Markov chain, we used a variety of moves (described in detail below) that operate by adding, removing, replacing, or modifying ETs. We call those ETs whose operation has $k$ offspring $k$-ET. For example, $x + a$ is a 2-ET (because the sum has two offspring: $x$ and $a$), $\sin x$ is a 1-ET (because the sine function has a single offspring: $x$), and $x$ is a 0-ET. Expressions such as $\sin(x + a)$ are not ETs because they contain more than one operation.

We designed an MCMC algorithm to sample expressions from the posterior distribution $p(f_i | D)$, which gives the plausibility of an expression given the observed data. This distribution is given by Eq. 3. We used three types of move to update mathematical expressions (Fig. 1):

1) Node replacement. We replaced a node in the expression tree (selected uniformly at random among all nodes in the tree) by a randomly selected node, with the only restriction that the replacement must have the same number of offspring as the original node (for example, a "+" node can be replaced by a "*" node but not by an "exp" node). The offspring branches remain unchanged.

2) Root addition (RA). We added a new root to the expression tree. The new root can be any operation. If the operation takes one offspring, then the old expression tree becomes the offspring of the new root; otherwise, the old expression tree becomes the leftmost offspring of the operation, with the other offspring being randomly chosen 0-ETs (that is, variables or parameters). To be more precise, at the beginning of a sampling process, all possible replacement roots are enumerated (all operations with all possible 0-ET offspring in all positions other than the leftmost branch, which is left empty); when a RA is attempted, the new root is chosen uniformly among this list of candidates. The reverse move, root removal (RR), consists of removing the root of the expression tree and all its offspring except for the leftmost branch, which becomes the new expression tree. RR is only possible when all branches except the leftmost one are 0-ETs.

3) Elementary tree replacement (ETR). We replaced a randomly selected ET in the complete expression tree by another randomly selected ET. At the beginning of a sampling process, all possible ETs are enumerated (all operations with all possible 0-ET offspring). In each move, an ET (chosen uniformly at random among all ETs in the expression tree) is replaced by an ET chosen uniformly at random among all possible ETs.

The ETR move introduces small variations to expression trees and is therefore the major source of expression variation. By contrast, node replacement moves often introduce major changes in expressions (for example, replacing a sum by a product often alters a model very significantly) and are therefore not very efficient. However, they are useful in that they represent long jumps in the space of models. Last, root replacement is the only move that can make trees grow/shrink at the top and is useful to add/remove terms to models that are already reasonable. Without root replacement, adding an additive term to an expression tree would require disassembling the whole tree and assembling it again with the additional term from the beginning.

### MCMC acceptance rules
At each MCMC step, we attempt one of the three moves described above and accept or reject the proposed move according to Metropolis' rule (36)

$$p_{\text{accept}}(f_i \to f_f) = \min\left\{1, \frac{p(f_f | D)g(f_i | f_f)}{p(f_i | D)g(f_f | f_i)}\right\} \quad (5)$$

where $g(f_f | f_i)$ is the distribution of movement proposal $f_i \to f_f$ (where f and i stand for final and initial, respectively). This rule ensures that the stationary distribution is $p(f_i | D)$

$$p(f_i | D) = \frac{1}{Z} \exp\left[-\mathcal{L}(f_i)\right] \quad (6)$$

with $Z$ being a normalizing constant and $\mathcal{L}(f_i)$ being the description length, as defined and approximated in the main text

$$\mathcal{L}(f_i) \approx \frac{B(f_i)}{2} - \log p(f_i) \quad (7)$$

For all samples in the paper, we attempt a root replacement move 5% of the time, a node replacement move 45% of the time, and an ETR 50% of the time. These rates are arbitrary and should not affect the achievement or the equilibrium distribution but do have an effect on how fast the Markov chain converges to the equilibrium distribution.

In what follows, we describe the specific form of the acceptance (Eq. 5) rule for each movement.

1) Node replacement (NR). Since the node to change is chosen uniformly at random and so is the new node, the move is symmetric, that is, the probability of attempting a move and its reverse are the same $g(f_f|f_i) = g(f_i|f_f)$. Therefore, the acceptance probability in this case is given simply by the difference in the description lengths of the expression trees

$$p_{accept}^{NR} = \min\{1, \exp[-\Delta \mathcal{L}]\} \quad (8)$$

2) RA and RR. In this case, the proposal distribution for the RA and RR movements is not symmetric. The RR move is deterministic in that it always affects the existing root of the expression tree. By contrast, the RA move involves selecting uniformly at random among all possible $N_{root}$ roots that can be added (which are enumerated once at the beginning of the sampling process, as described above). Therefore, if $p_{RR}$ is the probability of selecting this type of move, then $g(RR) = p_{RR}$ and $g(RA) = p_{RR}/N_{root}$ so that $g(RR)/g(RA) = N_{root}$, which gives the following acceptance rules

$$p_{accept}^{RA} = \min\{1, N_{root} \exp[-\Delta \mathcal{L}]\} \quad (9)$$

$$p_{accept}^{RR} = \min\left\{1, \frac{\exp[-\Delta \mathcal{L}]}{N_{root}}\right\} \quad (10)$$

where $N_{root}$ is the number of possible roots among which we choose in the RA move.

3) ETR. The ETR move is the most involved in terms of defining the move proposal probabilities necessary to define the acceptance rule. First, we need to specify exactly how we choose the attempted move. We start by selecting the orders $o_i$ and $o_f$ of the existing (initial, i) and replacement (final, f) ETs. For example, we may choose to replace a 0-ET ($o_i = 0$, a constant or a variable) by a 2-ET ($o_f = 2$, e.g., a sum or a product). In this initial choice, we need to take into consideration the number $n_{if}$ of options that we have to choose $o_i$ and $o_f$ (for example, if a tree has reached the maximum allowed size $o_f \leq o_i$ necessarily). Once the orders $o_i$ and $o_f$ are selected, we need to account for (i) all the possible ETs of order $o_i$ in the initial tree $\Omega_i$ that we can choose to replace and (ii) all the possible ETs of order $o_f$ that we can choose to replace the ET in the initial tree, $s_f$. If $p_{ETR}$ is the probability with which we choose this move, then taking into account the previous considerations $g(o_f|o_i) = p_{ETR} \times 1/n_{if} \times 1/\Omega_i \times 1/s_f$. Conversely, $g(o_i|o_f) = p_{ETR} \times 1/n_{fi} \times 1/\Omega_f \times 1/s_i$, so that the acceptance rule is

$$p_{accept}^{ETR} = \min\left\{1, \frac{n_{if}\Omega_i s_f}{n_{fi}\Omega_f s_i} \exp[-\Delta \mathcal{L}]\right\} \quad (11)$$

To validate that these moves and these acceptance rules lead to sampling from the equilibrium distribution $p(f_i|D)$, we use the same example as in Fig. 1. We generate data as in Fig. 1 and sample expressions using MCMC, limiting the expressions to use only "+" and "sin" operations, a single variable $x$, a single parameter $a$, and a maximum of seven nodes. In fig. S1, we show that, as expected, the equilibrium distribution is $p(f_i|D)$.

### Avoidance of expression duplicates in MCMC

The mapping of expressions to expression trees is not one to one because several expression trees can represent the same expression (for example, the expression $x + a$ can be encoded in an expression tree with $x$ or with $a$ on the left branch). To avoid overcounting expressions, we internally reduce all expression trees to a "canonical form" using the Python module Sympy (37). Then, if an expression tree that is visited for the first time reduces to an expression that is equivalent to that of a previously visited expression tree, then the current tree is forbidden (assigned infinite description length) and never visited again, so that only one representative of each expression remains.

### Parallel tempering MCMC

The search space for the MCMC is extremely rugged, with some neighboring expressions having very different description lengths. This makes the sampling process problematic and prone to getting trapped in local minima. To partly alleviate this problem, we used parallel tempering (38). In traditional parallel tempering (typically used in physics for spin glasses and other disordered systems), several replicas of the sampling process are kept at logarithmically spaced and increasing temperatures—at high temperatures, the sampling easily escapes local minima, whereas at low temperatures, the sampling explores configurations that are physically more meaningful. From time to time, samples at consecutive temperatures are switched with a rule that guarantees detailed balance at all temperatures. Therefore, the lowest temperature sample is always equilibrated at the desired temperature, but the configuration space is explored more efficiently because of the high temperature samples exploring larger portions of the space.

For our purposes here, we introduce a computational temperature $T$ and sample from the distributions

$$p(f_i, T_k) = \frac{1}{Z} \exp\left[-\frac{B(f_i)}{2T_k} + \log p(f_i)\right] \quad (12)$$

With this definition, $p(f_i, T = 1) = p(f_i|D)$ is the equilibrium distribution from which we aim to sample, and $p(f_i, T = \infty) = p(f_i)$ is the prior distribution, independent of the data.

The results of the manuscript were obtained with 40 different temperatures defined as $T_k = 1.05^k$, with $k = \{0,1,…,39\}$. From all these, we take expressions only from the $T_0 = 1$ sample (Fig. 3). In addition, for the Nikuradse dataset, we used four restarts of the parallel tempering MCMC so that the sampling is even better equilibrated.

### Parsing of Wikipedia expressions for expression priors

We compiled all the mathematical expressions included in pages listed under Wikipedia's "List of scientific equations named after people," which we surmise often correspond to phenomenological models of the kind that we aim to identify (as opposed, for example, to derivations of mathematical identities, proofs of mathematical theorems, etc.).

In Wikipedia, expressions are encoded in such a way that some of them are ambiguous without the context provided by the text in the page, so fully automatic parsing of expressions is virtually impossible. For example, the expression $x^a + x^b$ could represent the sum of the two components of a vector or the sum of two powers of $x$. We designed an algorithm that parses Wikipedia expressions into Sympy (37), using heuristics for the ambiguous cases (the full code for expression parsing is available at https://bitbucket.org/rguimera/machine-scientist/src/no_degeneracy/Process-Formulas/). We then selected three random subsets of 200 expressions out of the 4080 expressions that could be parsed and verified manually that the parsings were meaningful (accepting parsings that may be wrong once the context is known but that are otherwise mathematically correct). We refined the heuristics until we found that there were fewer than 5% of errors in the parsed expressions of all three subsets. In table S2, we show a small sample of the corpus.

### Prior definition

To define the prior distribution over mathematical expressions, we took advantage of the fact that mathematical expressions can be represented as graphs and used an approach based on exponential random graph models (20–23). As in exponential random graphs models, we aimed to generate mathematical expressions (graphs) with statistical properties similar to those in the corpus. Specifically, we aimed to generate mathematical expressions for which the average number of each operation per expression is the same as in the corpus (for example, in the corpus, the average number of sums per expression is $\langle n_+ \rangle$ = 0.312). We also aimed to reproduce the average of the square of the number of each operation per expression (for example, the average of the square of the number of sums per expression, $\langle n_+^2 \rangle$ = 0.731) (23, 24). As in exponential random graph models, we selected the prior probability $p(f_i)$ that generates expressions with these desired average properties and that, at the same time, is maximally uninformative and therefore consistent with the maximum entropy principle (14, 23, 39). This prior is given by

$$p(f_i) = \sum_{o \in O} [\alpha_o n_o(f_i) + \beta_o n_o^2(f_i)] \quad (13)$$

where the sum is over all operations considered $O = \{+, \exp, \ldots\}$ (table S1), and $\alpha_o$ and $\beta_o$ are hyperparameters that we fitted so that expressions generated from the prior were consistent with the corpus (22, 23).

### Fitting of prior hyperparameters

To fit the $\alpha_o$ and $\beta_o$ hyperparameters, we proceeded as follows:

1) Obtain, from the empirical corpus of mathematical expressions, the average number $\langle n_o \rangle^{\text{target}}$ of each operation $o$ per expression and the average of the square $\langle n_o^2 \rangle^{\text{target}}$ of the number of each operation per expression.

2) Set some initial values for the parameters.

3) Repeat until the change in parameters values is small enough:

   (i) Generate, using the MCMC, a large number of expressions (typically 1 million to 10 million).
   
   (ii) Measure $\langle n_o \rangle^{\text{measured}}$ and $\langle n_o^2 \rangle^{\text{measured}}$ in the generated formulas.
   
   (iii) Update the $\alpha_o$ parameters as follows

$$\alpha_o \leftarrow \alpha_o + \varepsilon \lambda \frac{\langle n_o \rangle^{\text{measured}} - \langle n_o \rangle^{\text{target}}}{\langle n_o \rangle^{\text{target}}} \quad (14)$$

where $\lambda$ is a fixed parameter (typically between 0.01 and 0.05) and $\varepsilon$ is a random number in [0, 1] generated independently for each update.

   (iv) Update the $\beta_o$ parameters as follows

$$\beta_o \leftarrow \beta_o + \varepsilon \lambda \frac{\langle n_o^2 \rangle^{\text{measured}} - \langle n_o^2 \rangle^{\text{target}}}{\langle n_o^2 \rangle^{\text{target}}} \quad (15)$$

where $\lambda$ is a fixed parameter (typically between 0.01 and 0.05) and $\varepsilon$ is a random number in [0,1] generated independently for each update. If one $\beta_o$ becomes negative, then its value is set to 0.

   (v) Repeat from (i).

This method is adapted from an exisiting method to fit the parameters of exponential random graphs (22). A typical evolution of the parameter values is shown in fig. S2, and the parameter values obtained are listed in tables S3 and S4 (Supplementary text S2).

### Benchmark machine scientists

We benchmarked the performance of the Bayesian machine scientist against three other machine scientists: Eureqa (5), EPLEX (6), and EFS (13). Eureqa uses a genetic algorithm to search the space of expressions (5). Eureqa requires that a complexity penalty is set for each operation type; we set these penalties to their default values and selected the same basic operations used by the Bayesian machine scientist. We also selected the default fitness function. We ran Eureqa for several weeks in each experiment, until the number of evaluated expressions is, at least, $10^{13}$. Eureqa is available at www.nutonian.com/download/eureqa-desktop-download/.

EPLEX is another algorithm based on genetic programming and was the top performer in a recent systematic comparison of state of the art machine scientists (4). We ran EPLEX in the same conditions as in that study (implementation available at https://epistasislab.github.io/ellyn/); in particular, we obtained $10^6$ models with a population size of 1000 and 1000 generations. Moreover, we repeated this procedure 10 times and selected the realization with the best fitness among the 10 repetitions. However, in all our experiments, EPLEX gives results that are considerably worse than Eureqa's, so we do not show its results in the main text (Supplementary text S7 and fig. S10).

Last, EFS is a method based on sparse regression that generates basis functions automatically using a genetic algorithm (13). Thus, this approach has the advantages of sparse regression, and, at the same time, it does not require a priori knowledge of the basis functions. EFS is an evolution of multiple regression genetic programming (12), which was also a top performer in (4), although somehow more inconsistent than EPLEX. EFS is fast and gives expressions within seconds. Thus, we were able to repeat the training process 100 times for each experiment and select the model with the best default fitness measure (mean squared error). We used the implementation of EFS available at http://flexgp.github.io/efs/.

### Gaussian process models

For the comparison of the out-of-sample predictions of Gaussian process models to those of the machine scientist, we proceeded as follows. We trained the Gaussian process using the scikit-learn implementation (40). We tested different kernels (including radial-basis function kernels, Matérn kernels, white kernels, rational quadratic kernels, exponential kernels, and linear and quadratic combinations of these), and we also tested the effects of using $D/r$ or its logarithm as the feature. Among these, we selected the combination that gave the best results (an radial-basis function plus white kernel with logarithmic $D/r$)—all results reported in the main text correspond to this combination.

### SUPPLEMENTARY MATERIALS

Supplementary material for this article is available at http://advances.sciencemag.org/cgi/content/full/6/5/eaav6971/DC1

Supplementary Text S1. Calculation of the BIC
Supplementary Text S2. Prior parameter values

**Acknowledgments:** We thank E. G. Altmann, L. A. N. Amaral, A. Arenas, J. Bonet Avalos, and D. Shasha for helpful comments and suggestions. We thank I. Arnaldo for help with the EFS software. We thank A. Arenas for pointing us toward the Nikuradse dataset. We thank M. De Domenico and A. Arenas for sharing the financial success dataset. We thank E. Bazellières and X. Trepat for sharing the cell adhesion dataset. **Funding:** This project has received funding from the Spanish Ministerio de Economia y Competitividad (FIS2015-71563-ERC, FIS2016-78904-C3-P-1, and DPI2016-75791-C2-1-P). F.A.M. acknowledges financial support by the Spanish MINECO grant PTQ-14-06718 (2016-2019) of the Torres Quevedo Programme. **Author contributions:** R.G. conceived the research. R.G., I.R., A.A.-M., F.A.M., M.M., and M.S.-P. contributed methods, wrote code for the computational experiments, and carried out computational experiments. All authors designed the computational experiments and interpreted the results, and wrote and edited the manuscript. **Competing interests:** The authors declare that they have no competing interests. **Data and materials availability:** A Python implementation of the Bayesian machine scientist and all data needed to evaluate the conclusions in the paper are publicly available from Bitbucket at https://bitbucket.org/rguimera/machine-scientist. Additional data related to this paper may be requested from the authors.

Submitted 11 October 2018
Accepted 20 November 2019
Published 31 January 2020
10.1126/sciadv.aav6971

**Citation:** R. Guimerà, I. Reichardt, A. Aguilar-Mogas, F. A. Massucci, M. Miranda, J. Pallarès, M. Sales-Pardo, A Bayesian machine scientist to aid in the solution of challenging scientific problems. *Sci. Adv.* **6**, eaav6971 (2020).